\documentclass[times,twocolumn,final,authoryear]{elsarticle}

\usepackage{ycviu}
\usepackage{framed,multirow}
\usepackage[T2A,T1]{fontenc}
\usepackage[utf8]{inputenc}
\usepackage[russian,english]{babel}
\usepackage{booktabs}
\usepackage{amsmath}

\usepackage{amssymb}
\usepackage{latexsym}

\usepackage{url}
\usepackage{xcolor}
\usepackage{makecell}
\usepackage{ulem}

\graphicspath{{./imgs/}}

\definecolor{newcolor}{rgb}{.8,.349,.1}

\journal{Computer Vision and Image Understanding}

\begin{document}

\ifpreprint
  \setcounter{page}{1}
\else
  \setcounter{page}{1}
\fi

\begin{frontmatter}

\title{The MSR-Video to Text Dataset with Clean Annotations}

\author[1,2,3]{Haoran \snm{Chen}} 
\author[1,2,3]{Jianmin \snm{Li}}
\author[5]{Simone \snm{Frintrop}}
\author[1,2,3,4]{Xiaolin \snm{Hu}\corref{cor1}} 
\cortext[cor1]{Corresponding author:  }
\ead{xlhu@tsinghua.edu.cn}

\address[1]{State Key Laboratory of Intelligent Technology and Systems, Department of Computer Science and Technology, Tsinghua University, Beijing, 100084, China}
\address[2]{Beijing National Research Center for Information Science and Technology, Tsinghua University, Beijing, 100084, China}
\address[3]{Institute for Artificial Intelligence, Tsinghua University, Beijing, 100084, China}
\address[4]{Chinese Institute for Brain Research (CIBR), Beijing, 100010, China}
\address[5]{Department of Informatics, University of Hamburg, Hamburg, 20148, Germany}

\received{1 April 2022}

\begin{abstract}
Video captioning automatically generates short descriptions of the video content, usually in form of a single sentence. 
Many methods have been proposed for solving this task.  
A large dataset called MSR Video to Text (MSR-VTT) is often used as the benchmark dataset for testing the performance of the methods.
However, we found that the human annotations, i.e., the descriptions of video contents in the dataset are quite noisy, e.g., there are many duplicate captions and many captions contain grammatical problems. 
These problems may pose difficulties to video captioning models for learning underlying patterns.
We cleaned the MSR-VTT annotations by removing these problems, then tested several typical video captioning models on the cleaned dataset.   
Experimental results showed that data cleaning boosted the performances of the models measured by popular quantitative metrics.
We recruited subjects to evaluate the results of a model trained on the original and cleaned datasets. 
The human behavior experiment demonstrated that trained on the cleaned dataset, the model generated captions that were more coherent and more relevant to the contents of the video clips. 
\end{abstract}


\end{frontmatter}


\section{Introduction}\label{sec:introduction}

\begin{figure}[h]
    \centering
    \includegraphics[width=0.9\linewidth]{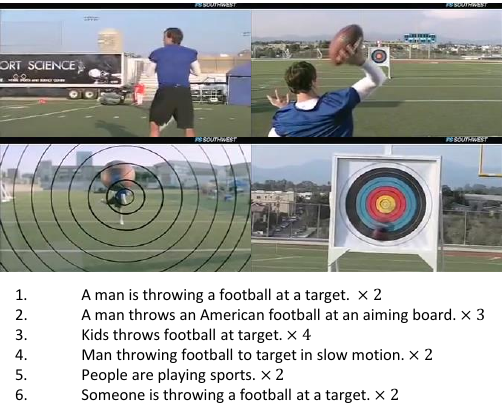}
    \caption{An example video clip (No. 4290, starting from 0) with duplicate annotations. $\times t$ denotes repeating  $t$ times}
    \label{fig:repeatedMSRVTT}
\end{figure}

The goal of video captioning is to summarize the content of a video clip by a single sentence,
which is an extension of image captioning \citep{cho2014learning, rennie2017self, yu2018topic, anderson2018bottom}. 
To accomplish it, one must use both computer vision (CV) techniques and natural language processing (NLP) techniques.
A benchmark dataset, called MSR-Video to Text 1.0 (MSR-VTT v1)\citep{Xu_2016_CVPR}, was released in 2016.
It contains 10,000 video clips and each clip is described by 20 captions,
which are supposed to be different, given by human annotators.
The dataset has become popular in the field of both video captioning and retrieval. 
Until March 31st, 2022, that work \citep{Xu_2016_CVPR} has been cited by 793 times according to Google scholar.

\begin{figure*}[h]
    \centering
    \includegraphics[width=0.7\linewidth]{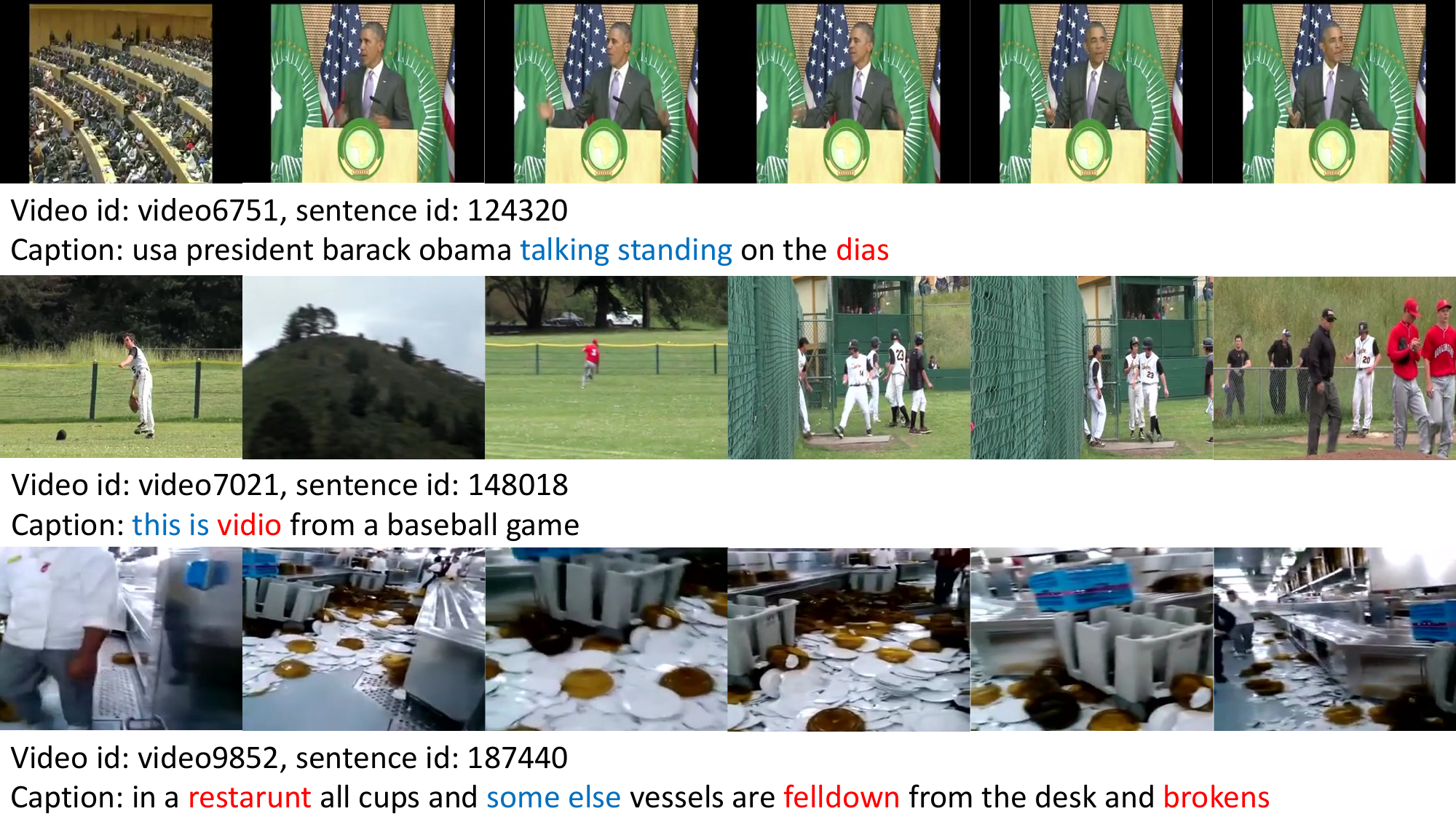}
    \caption{Three examples in the MSR-VTT dataset. The words in blue and red denote grammatical mistakes and spelling mistakes, respectively }
    \label{fig:defectsMSRVTT}
\end{figure*}

However, with a quick look, one can find many duplicate annotations, spelling mistakes and syntax errors in the annotations (Figs. \ref{fig:repeatedMSRVTT}, \ref{fig:defectsMSRVTT}).
It is unknown how many mistakes there are exactly in the dataset and whether/how these mistakes would influence the performance of the video captioning models.

We quantitatively analyzed the annotations in the MSR-VTT dataset, and identified four main types of problems.
First, thousands of annotations have duplicates for some of the video clips in the dataset.
Second, thousands of special characters, such as "+", "-", ".", "/", ":", exist in the annotations.
Third, thousands of spelling mistakes exist in the annotations.
Fourth, hundreds of sentences are redundant or incomplete.
We developed some techniques for cleaning the annotations to solve these problems.
Our experiments demonstrated that existing models of video captioning, trained on the cleaned training set,
had better performances compared to the results obtained by the models trained on the original training set.
A human evaluation study also showed that a state-of-the-art model trained on the cleaned training set generated better captions
than trained on the original training set in terms of semantic relevance and sentence coherence. 

The cleaned dataset will be made available on request.

\section{Related Work}

\subsection{Datasets}
Three datasets MSVD (also called YouTube2Text), MSR-VTT and VATEX, unlimited to a specific domain, are widely used in recent studies of video captioning as benchmarks and video retrieval as well.

MSVD was published in 2013 \citep{Guadarrama_2013_ICCV}. It contains 1970 video clips and roughly 80,000 captions.
Each video clip pairs with 40 captions.

MSR-VTT v1 was published in 2016 \citep{Xu_2016_CVPR}. It contains 10,000 video clips and 200,000 captions.
Each video clip pairs with 20 captions.
The MSR-VTT v2 dataset was proposed in the second Video Captioning Competition\footnote{Competition Website: \url{http://ms-multimedia-challenge.com/2017/challenge}}
using the MSR-VTT v1 dataset as the training and validation sets and additional 3000 video clips as the test set.
However, the annotations of the test set are not open to the public.

In 2019, a new large-scale video description dataset, named VATEX was presented, which is multilingual, linguistically complex and diverse in terms of video and annotations. It contains over 41,000 videos, reused from Kinetics-600, with 10 English text sentences for each of them \citep{9010676}.

\subsection{Recent Advances in Video Captioning}
Kinds of models or methods or algorithms have been proposed for video captioning.
With semantic concepts detected from the video, the probability distribution of each tag is integrated into the parameters of a recurrent unit in SCN \citep{DBLP:conf/cvpr/GanGHPTGCD17}.
Video captioning is improved by sharing knowledge with two related tasks on the encoder and the decoder of a sequence-to-sequence model \citep{DBLP:conf/acl/PasunuruB17}.
Reinforced learning is enhanced for video captioning with the mixed-loss function and the CIDEr-entailment reward in CIDEnt-RL \citep{pasunuru-bansal-2017-reinforced}.
Multiple modalities are fused by hierarchical attention, which helps to improve the model performance, in HATT \citep{DBLP:journals/ijon/WuWCSSW18}.
The video feature produced by Efficient Convolutional Network is fed into a video captioning model, which boosts the quality of the generated caption, in the model named ECO \citep{zolfaghari2018eco}.
In the GRU-EVE, the Short Fourier Transform is applied to video features and high level semantics is derived from the object detector in order to generate captions rich in semantics \citep{DBLP:conf/cvpr/AafaqALGM19}.
A memory structure is used to capture the comprehensive visual information across the whole training set for a word in the MARN \citep{Pei_2019_CVPR}.
The encoder employs a sibling (dual-branch) architecture to encode video clips in the SibNet \citep{liu-sheng-2018-sibnet}.
HACA fuses both global and local temporal dynamics existing in a video clip and generates an accurate description with knowledge from different modalities \citep{wang2018watch}.
Different expert modules are trained to provide knowledge for describing out-of-domain video clips in the TAMoE \citep{wang2019learning}.
The model called SAM-SS is trained under the self-teaching manner to reduce the gap between the training and the test phase with meaningful semantic features \citep{10.3389/frobt.2020.475767}.
Different types of representations are encoded and fused by the cross-gating block and captions are generated with Part-of-Speech information in the POS\_RL \citep{Wang_2019_ICCV}.
In the VNS-GRU, ``absolute equalitarianism'' in the training process is alleviated by professional learning while a comprehensive selection method is used to choose the best checkpoint for the final test \citep{chen2020delving}.
A new paradigm, named Open-book Video Captioning \citep{zhang2021open}, is adopted to generate natural language under the prompts of video-content-relevant sentences, unlimited to the video itself.

\section{Analysis and Cleaning of the MSR-VTT dataset}

Since MSR-VTT v2 uses MSR-VTT v1 for training and validation, and the annotations of the test set of MSR-VTT v2 are not open to the public,
we performed analysis on  MSR-VTT v1.

The MSR-VTT v1 dataset contains 10,000 video clips.
Its training set has 6,513 video clips,
the validation set has 497 video clips and
the test set has 2,990 video clips.
All clips are categorized into 20 classes with diverse contents and sceneries.
A total of 0.2 million human annotations were collected to describe those video clips. The training/validation/test sets have
130,260/9,940/59,800 annotations, respectively.
The vocabulary sizes of the training/validation/test set are 23,666/5,993/16,001, respectively.

\subsection{Special Characters}\label{step1}
There are 60 different characters in the dataset, including 0-9,
a-z and 24 special characters in Table \ref{table:special-character} (space is neglected).
Generally speaking, those special characters are not used to train a model.
We are intended to remove special characters while preserve information integrity in annotations.

We processed those special characters as follow:
\begin{enumerate}
    \item Some special characters were removed from the sentences, include "\#", "*", "+", ".", ":", "=", "\textgreater", "[", "]", "(", ")", "\textbackslash",
    where "[", "]", "(" and ")" were removed only when they were not in pairs.
    \item The contents between bracket pairs "()" and "[]" were removed.
    \item Special characters "-", "\textbar", "`", "@", "\_", "’", "/" were replaced with spaces.
    \item Characters from another language were replaced by the most similar English characters.
        For example, "\'{e}" was replaced by "e" in "\'{e}rror"
        and "\foreignlanguage{russian}{в}" by "b" in "\foreignlanguage{russian}{в}eautiful".
    \item "\&" between two different words was substituted by "and".
\end{enumerate}
In total, 7,248 out of 200,000 sentences and 4,190 out of 10,000 video clips  were corrected (Table \ref{table:numAbnorm}).

\begin{table}[htb]
    \begin{center}
                \caption{Special characters in the MSR-VTT dataset} \label{table:special-character}
        \begin{minipage}{0.8\linewidth}
            \begin{tabular*}{\linewidth}{@{\extracolsep{\fill}}cccccc@{\extracolsep{\fill}}}
                \toprule
                    \#			& \$  			& \%  		& \&  		& (   								& ) \\
                    \midrule
                    *   			& +   		& -   			& .   			& $\_$   							& : \\
                    \midrule
                    = 			& \textgreater   	& @   		& [     		& \textbackslash  					& ] \\
                    \midrule
                    / 			& `   			& \textbar   	& \'{e} 		& \foreignlanguage{russian}{в} 			& ’ \\
                \bottomrule
            \end{tabular*}
        \end{minipage}
    \end{center}
\end{table}

\subsection{Spelling Mistakes}\label{step2}

Massive spelling mistakes were found in the annotations during manual check.
Tokenization is a process of demarcating a string of an input sentence into a list of words.
After tokenization on each of the sentences, we used  a popular spelling check software Hunspell \footnote{Available at \url{https://hunspell.github.io}} to check spelling errors.

Before Hunspell was used to do spelling checks, 784 new words were added to its vocabulary.
These words were chosen manually by four criteria:
\begin{enumerate}
    \item word abbreviations that are popular, eg. F1, WWF, RPG;
    \item specific terms that are widely used, eg. Minecraft, Spongebob, Legos;
    \item new words that are popular on the Internet, eg. gameplay, spiderman, talkshow;
    \item names of persons, eg. Mariah, Fallon, Avril.
\end{enumerate}
	
After that, spelling mistakes were found in 19,038 annotations out of 200,000 annotations.
21,826 words might have spelling mistakes suggested by Hunspell.
Those candidates were corrected in the following steps:
\begin{enumerate}\label{list:correct}
    \item Substituted British English spellings with the corresponding American English spellings. For instance, colour $\rightarrow$ color, travelling $\rightarrow$ traveling, programme $\rightarrow$ program, practising $\rightarrow$ practicing, theatre $\rightarrow$ theater. There were 61 such pairs.
    \item Split unusual words that were created by concatenating two different words,
    e.g. rockclimbing $\rightarrow$ rock climbing, blowdrying $\rightarrow$ blow drying,
    swordfighting $\rightarrow$ sword fighting, screencaster $\rightarrow$ screen caster, rollercoaster $\rightarrow$ roller coaster.
    In total, 34 distinct words were found.
    \item Corrected words that truly contain spelling mistakes, e.g., discusing $\rightarrow$ discussing, explaning $\rightarrow$ explaining, coversation $\rightarrow$ conversation, vedio $\rightarrow$ video, diffrent $\rightarrow$ different.
\end{enumerate}
In total 35,668 words were substituted, split or corrected in these three steps for 27,954 sentences in 7,829 video clips as shown in Table \ref{table:numAbnorm}.

\subsection{Duplicate Annotations}\label{step3}
Duplicate sentences were discovered in many annotations of video clips (Fig. \ref{fig:repeatedMSRVTT}).
For each video clip, duplicates were removed.
The similarity between two sentences was defined as follow
\begin{equation}\label{eq: similarity}
    s_{a,b} = 0.5 (\frac{\mu{(\mathbf{a}, \mathbf{b})}}{\iota{(\mathbf{a})}} + \frac{\mu{(\mathbf{a}, \mathbf{b})}}{\iota{(\mathbf{b})}}),
\end{equation}
where $\iota{(\mathbf{x})}$ denotes the word count in the sentence $\mathbf{x}= \{x_1, x_2, ...\}$
and $\mu{(\mathbf{a}, \mathbf{b})}$ denotes the word count of the longest common subsequence in $\mathbf{a}$ and $\mathbf{b}$.
$\mu{(\mathbf{a}, \mathbf{b})}$ is defined as follows,
\begin{align}
    \mu{(\mathbf{a}, \mathbf{b})} = \max{(\iota{(\mathbf{c})})} \nonumber \\
     \qquad \mathrm{s.t.}\quad \mathbf{c} \in \mathbf{a}, \mathbf{c} \in \mathbf{b},
\end{align}
where $\mathbf{c} \in \mathbf{a}$ stands for that $\mathbf{c}$ is a subsequence of $\mathbf{a}$.
Word $w_1$ and word $w_2$ were regarded as the same if the Levenshtein distance \citep{levenshtein1966binary} between them was less than or equal to $\bar{e}$.
Two sentences were regarded as duplicated if $s_{a,b} > \bar{s}$, where $\bar{s}$ is the similarity threshold.
With proper values of $\bar{e}$ and $\bar{s}$, we could find duplicated sentences that had little difference.
For example, considering the second pair of sentences in Table \ref{table:duplicates},
the character ``m'' is missing in the word ``woan'' and the second sentence just has one more word ``young'' than the first sentence.
These two sentences are almost the same in terms of meaning.

\begin{table*}[htb]
    \begin{center}
    	\begin{minipage}{0.8\textwidth}
            \caption{Examples of duplicates and its similarity value} \label{table:duplicates}
            \begin{tabular*}{\textwidth}{@{\extracolsep{\fill}}ll@{\extracolsep{\fill}}}
                \toprule
                Sentences 					 							                    	& Similarity\footnotemark[1]								\\
                \midrule
                a woman is walking down the aisle in a wedding 		    						& \multirow{2}*{0.86, 0.96, 0.96}							\\
                a woman is walking down the isle in a wedding dress	  							& 													\\ 
                \midrule
                a man is talking to a woan  								          			& \multirow{2}*{0.80, 0.94, 0.94}							\\
                a young man is talking to a woman						          				& 													\\ 
                \midrule
                a woman is singing on a music video					          					& \multirow{2}*{0.83, 0.94, 0.94}							\\
                a young woman is singing in a music video				      					& 													\\
                \bottomrule		
            \end{tabular*}
            \footnotetext[0]{Note: Each pair of sentences describe the same video clip.}
            \footnotetext[1]{ Those values are calculated by each pair of sentences with $\bar{e}=0,1,2$. }
        \end{minipage}
    \end{center}
\end{table*}

After duplicate removal, 183,856 video annotations remained in the dataset with 119,625 in the training set, 9,126 in the validation set and 55,105 in the test set with the hyper-parameters $\hat{e}=0, \hat{s}=0.85$, 
tuned in Section \ref{sec:exper}. 
In another word, 17,733 sentences were removed in 7,129 video clips (Table \ref{table:numAbnorm}).
Each clip has 9 annotations at least, 20 at most and 18.4 on average.

\begin{figure}[htb]
    \centering
      \begin{tabular} {c}
        \toprule
          \begin{minipage}{\linewidth}\includegraphics[width=\textwidth]{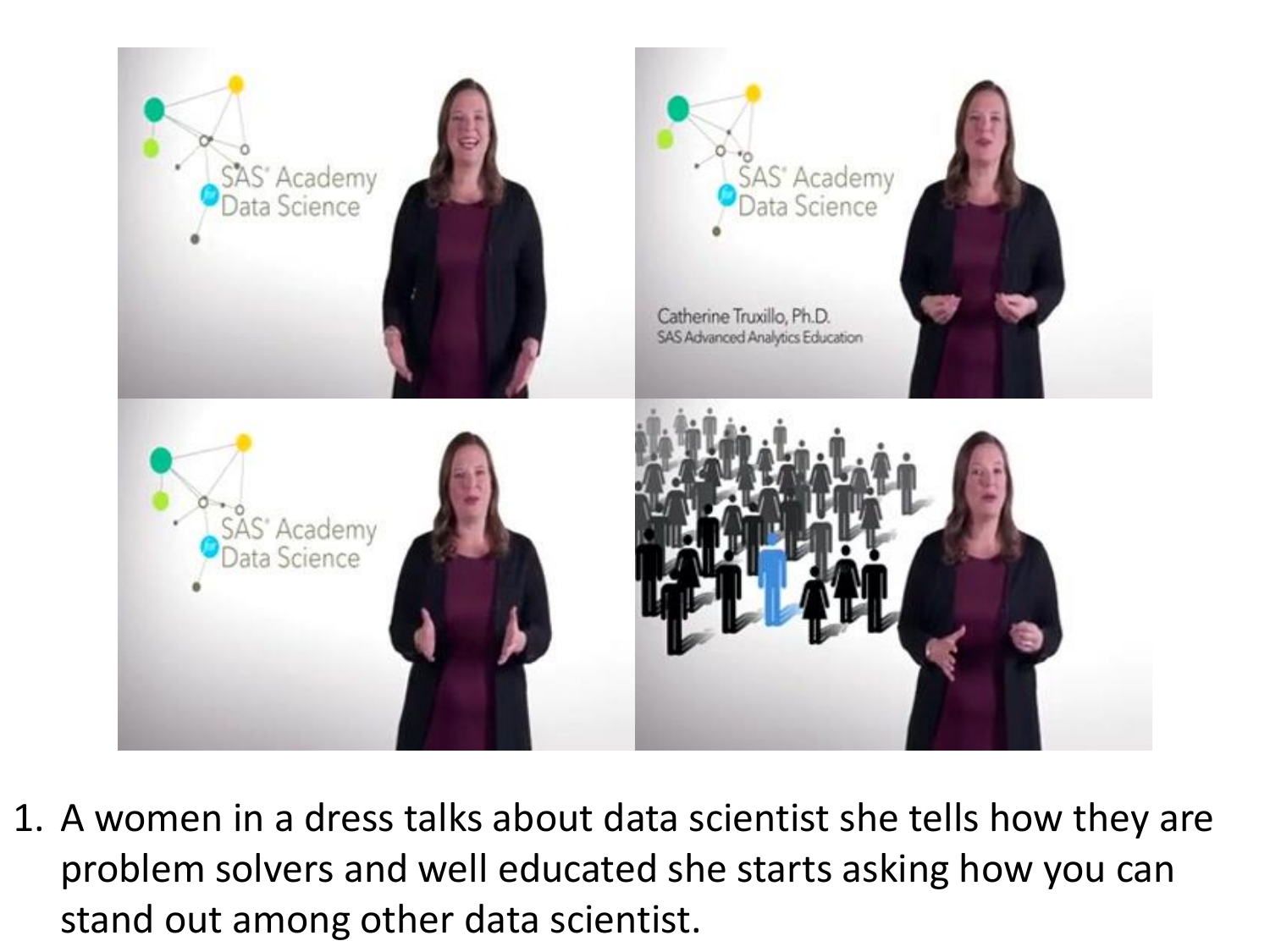}\end{minipage} \\
          \begin{minipage}{\linewidth}\includegraphics[width=\textwidth]{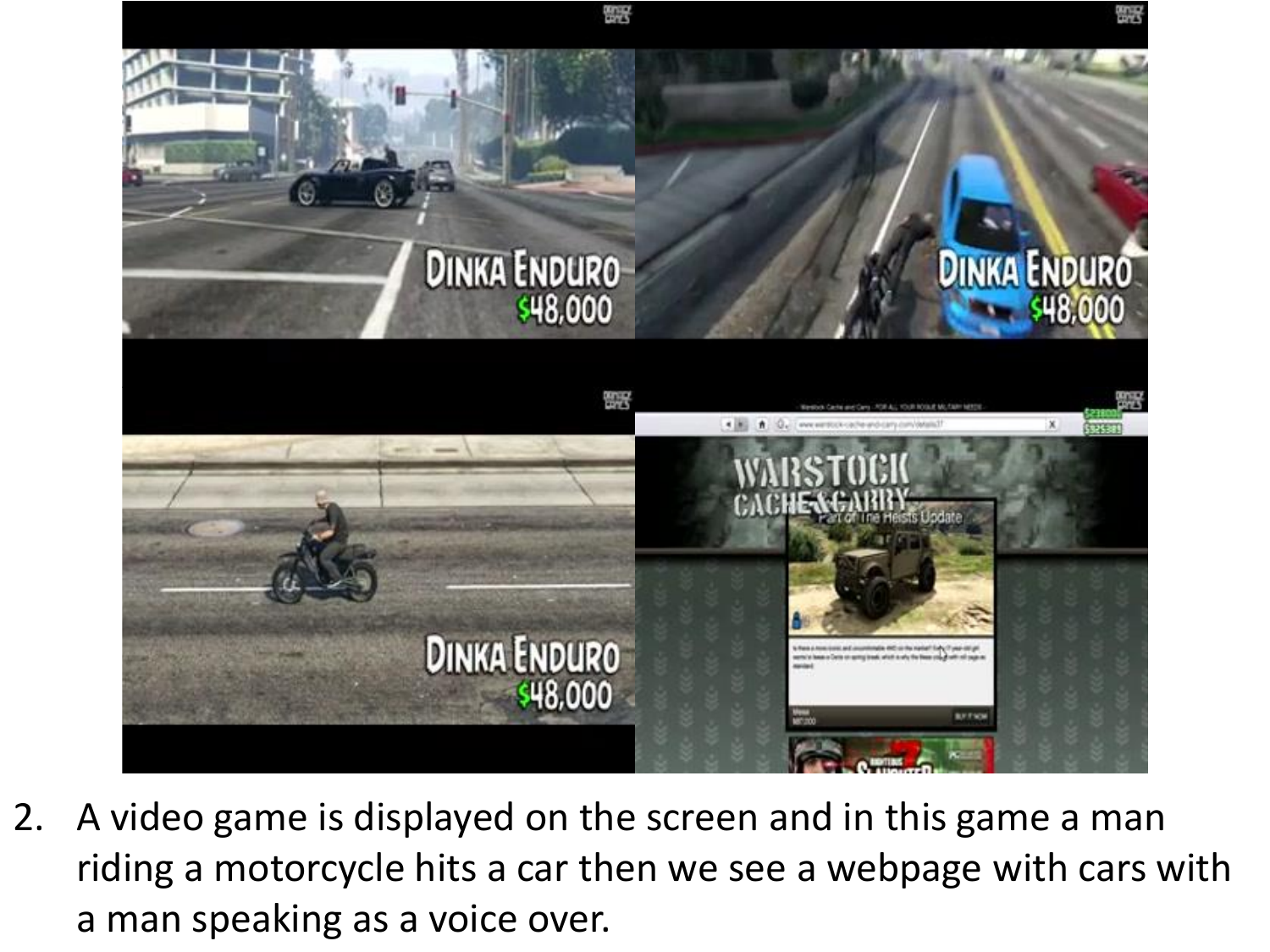}\end{minipage} \\
        \bottomrule
      \end{tabular}
    \caption{Redundancy samples in the MSR-VTT dataset. Caption 1 can be divided into three sentences. And Caption 2 can be divided into two or three sentences}
    \label{figure:msr-vtt-redundant-sentences}
\end{figure}

\subsection{Successive Sentences without Punctuations}\label{step4}
In the task of video captioning, we expect each annotation contains one sentence.
For many annotations in the dataset, each of them consists of multiple sentences.
In Fig. \ref{figure:msr-vtt-redundant-sentences}, the first annotation can be split into three complete sentences:
"A women in a dress talks about data scientist.",
"She tells how they are problem solvers and well educated.", 
"She starts asking how you can stand out among other data scientist."
It causes two potential problems.
First, the models trained on such annotations may output grammatically problematic sentences because these annotations are syntactically incorrect.
Second, such annotations in the test set are no longer reliable ground truth so that the metrics, computed with them, are not reliable, neither.

\begin{table}[htb]
    \begin{center}
    	\begin{minipage}{\linewidth}
            \caption{Impact of each step in terms of sentences and videos} \label{table:numAbnorm}
            \begin{tabular*}{\textwidth}{@{\extracolsep{\fill}}lcc@{\extracolsep{\fill}}}
                \toprule
                Step									& $Sentence_c$\footnote[1]{ The number of sentence that are corrected in a step }	& $Video_c$\footnote[2]{ The number of video that are corrected in a step }			\\
                \midrule
                Special Characters (Section \ref{step1})		& 7,248						& 4,190							\\
                Spelling Mistakes (Section \ref{step2})		& 27,954						& 7,829							\\ 
                Duplicate Annotations (Section \ref{step3})		& 17,733						& 7,129							\\
                Successive Sentences (Section \ref{step4})	& 5,543						& 2,758							\\ 
                \bottomrule		
            \end{tabular*}
        \end{minipage}
    \end{center}
\end{table}

However, many annotations in the dataset consist of multiple sentences. 
To solve it, one needs to manually separate the annotations into several complete sentences and merge them into a single sentence.
For the sake of efficiency, the annotations, which consist of several sentences, were only divided and merged for the test set. 
For the training and validation sets, the sentences longer than $l_a+2\sigma$, where $l_a$ and $\sigma$ denote the average sentence length and its standard deviation, were truncated respectively. 
After the process on it, 5,543 sentences were corrected in 2,758 video clips (Table \ref{table:numAbnorm}).

Table \ref{table:origincleaned} contains some random samples from original captions and cleaned ones. 
As shown in it, some most obvious mistakes are corrected and some redundant words are deleted.

    \begin{table*}[htb]
        \begin{center}
        	\begin{minipage}[c]{\linewidth}
                \caption{ Comparison of random samples from original captions and cleaned captions }
                \label{table:origincleaned}
                \begin{tabular*}{\textwidth}{@{\extracolsep{\fill}}cl@{\extracolsep{\fill}}}
                \toprule
                Sen Id			&	Sentence																						\\
                \midrule
                51307			& \makecell[l]{Animated hedgehog {\color{red}{complainging}} about being bored and a flying bug introduces sonic and the secret rings \\extreme party \sout{games}}	\\
                cleaned			& \makecell[l]{Animated hedgehog {\color{blue}complaining} about being bored and a flying bug introduces sonic and the secret rings \\extreme party} \\
                \midrule
                83933			& \makecell[l]{A man s hands are holding a {\color{red}red/orange} screwdriver and he shows u how to lock and \sout{unlock a deadbolted door}\\ \sout{with a key and a screwdriver while explaining his actions}}\\
                cleaned			& {a man s hands are holding a red orange screwdriver and he shows u how to lock and}\\
                \midrule
                188904			& \makecell[l]{An {\color{red}advertisment} to subscribe to {\color{red}weelious}}\\
                cleaned			& {An {\color{blue}advertisement} to subscribe to {\color{blue}rebellious}}\\
                \midrule
                57346			& \makecell[l]{A man is touching and talking about brake cables \sout{(and ziptying them/adding a pad)} the clutch and a handle for\\ what seems to \sout{be a motorcycle}}\\
                cleaned			& {A man is touching and talking about brake cables the clutch and a handle for what seems to}\\
                \midrule
                130327			& \makecell[l]{In a scene from a {\color{red}spanish-speaking} film a man breaks through a wooden door and confronts several \sout{other men inside}}\\
                cleaned			& {In a scene from a {\color{blue}spanish speaking} film a man breaks through a wooden door and confronts several}\\
                \midrule
                132787			& \makecell[l]{The girl is walked their {\color{red}warand} and she is giving flying {\color{red}kissshe} is {\color{red}weae} the pink \sout{topnear the green grass land}}\\
                cleaned			& {The girl is walked their {\color{blue}war and} and she is giving flying {\color{blue}kiss she} is {\color{blue}wear} the pink}\\
                \bottomrule		
                \end{tabular*}
                \footnotetext[1]{ {\color{red}Red} color denotes an error, {\color{blue}blue} color denotes modifications.}
            \end{minipage}
        \end{center}
    \end{table*}

\section{Experiments} \label{sec:exper}
Experiments were conducted on the original and cleaned MSR-VTT datasets with several existing video captioning models,
SCN \citep{DBLP:conf/cvpr/GanGHPTGCD17}, ECO \citep{zolfaghari2018eco}, SAM-SS \citep{10.3389/frobt.2020.475767} and VNS-GRU \citep{chen2020delving}.
They were trained for 30, 30, 50, 80 epochs, respectively.
They were evaluated on the validation set at the end of each epoch.
The first two models used the early stopping strategy with cross-entropy loss as the indicator.
The last two models used the Comprehensive Selection Method  to select a checkpoint for testing \citep{chen2020delving}.
For the sake of fair comparison, the experiment settings were the same as the original papers.
The two hyper-parameters $\bar{e}$ and $\bar{s}$ (see section \ref{step3}) were set to 0 and 0.85 in our experiments, unless otherwise stated.

\subsection{Evaluation Metrics}
BLEU, CIDEr, METEOR and ROUGE-L were adopted as objective metrics for evaluating the results of the models.
BLEU is a quick and easy-to-calculate metric, originally used for evaluating the performance of machine translation models \citep{papineni-etal-2002-bleu}.
CIDEr is a metric that captures human consensus \citep{cider-2015}.
METEOR is a metric that involves precision, recall and order correlation, based on unigram matches \citep{banerjee-lavie-2005-meteor}.
ROUGE-L is a metric that determines the quality of a summary by finding the longest common subsequence \citep{lin-2004-rouge}.
Besides these individual metrics, an overall score is presented to combine all of these metrics \citep{10.3389/frobt.2020.475767}:
\begin{equation}
	O_i=(\frac{B4_i}{B4_b} + \frac{C_i}{C_b} + \frac{M_i}{M_b} + \frac{R_i}{R_b}) / 4, \label{eq:overall}
\end{equation}
where the subscript $i$ denotes the model $i$ and the subscript $b$ denotes the best score of the metric $b$ over a group of models for comparison.
B4, C, M, R and O denote BLEU-4, CIDEr, METEOR, ROUGE-L and the overall score \eqref{eq:overall}, respectively.
\begin{table}[htb]
    \begin{center}
         \caption{Influence of Edit Distance Threshold $\bar{e}$ on the remaining annotation count and the performance of the model VNS-GRU}
    	\begin{minipage}{\linewidth}
            \label{table:edit distance}
            \begin{tabular*}{\textwidth}{@{\extracolsep{\fill}}ccccccc@{\extracolsep{\fill}}}
                \toprule
                $\bar{e}$        					& SC	\footnotemark[1]	& B4					& C 			  	& M 			  	& R  				& O 			\\
                \midrule
                $0$							& 184,078				& 47.6				& 52.6  			& 30.4			& 64.1			& 0.9988		\\
                $1$							& 183,856				& 47.2				& 52.2			& 30.2			& 64.1		  	& 0.9931		\\
                $2$							& 183,545				& 47.2				& 52.4			& 30.5			& 64.2			& 0.9969		\\
                \bottomrule		
            \end{tabular*}
            \footnotetext[0]{Note: $\bar{s}=0.9$. All metric values are presented in percentage.}
            \footnotetext[1]{SC represents the number of remaining sentences in the dataset.}
        \end{minipage}
    \end{center}
\end{table}
\subsection{Influence of Edit Distance Threshold and Similarity Threshold on Duplicates Removal}\label{experiment:removal}
In the step of removing duplicated annotations, there are two hyper-parameters: the edit distance threshold $\bar{e}$ and similarity threshold $\bar{s}$.
 The sensitivity of the hyper-parameters were investigated on the output of this step.
As shown in Table \ref{table:edit distance}, the threshold of edit distance $\bar{e}$ was inversely proportionate to the remained sentence count.
The performance of the model VNS-GRU was the best when $\bar{e}=0$.
As shown in Table \ref{table:similarity}, the threshold of similarity was proportionate to the remained sentence count.
The performance of the model VNS-GRU was the best when $\bar{s}=0.85$.
\begin{table}[htb]
    \begin{center}
         \caption{Influence of Similarity Threshold $\bar{s}$ on the remaining annotation count and the performance of the model VNS-GRU} \label{table:similarity}
    	\begin{minipage}{\linewidth}
            \begin{tabular*}{\textwidth}{@{\extracolsep{\fill}}ccccccc@{\extracolsep{\fill}}}
                \toprule
                $\bar{s}$ 					 	& SC\footnotemark[1] 		& B4					& C 			  	& M 			  	& R  			  	& O						\\
                \midrule
                $0.75$						& 175,539					& 46.8				& 53.6			& 30.4			& 64.0			& 0.9850					\\
                $0.80$						& 179,169					& 47.6				& 54.2			& 30.4			& 64.3			& 0.9933					\\
                $0.85$						& 182,264					& 47.4				& 55.0  			& 30.7			& 64.2			& 0.9982					\\
                $0.90$						& 183,705					& 47.4				& 53.7			& 30.4			& 64.0			& 0.9890					\\
                $0.95$						& 185,219					& 47.5				& 55.0			& 30.5			& 64.4			& 0.9978					\\
                $1.00$						& 185,330					& 47.6				& 53.4			& 30.2			& 63.9			& 0.9867					\\
                \bottomrule		
            \end{tabular*}
            \footnotetext{Note: $\bar{e}=0$.} 
            \footnotetext[1]{SC represents the number of remaining sentences in the dataset.}
	\end{minipage}
    \end{center}
\end{table}
Table \ref{table:duplicates} shows that with the method described in the Section \ref{step3}, we can find similar sentences, in terms of semantics, with one or two words different.

\begin{table*}[htb]
    \begin{center}
         \caption{ Results on the original/cleaned MSR-VTT dataset } \label{table:performanceMSRVTTNewOld}
    	\begin{minipage}{0.9\textwidth}
            \begin{tabular*}{\textwidth}{@{\extracolsep{\fill}}lccccc@{\extracolsep{\fill}}}
                \toprule
                Model                                   							&   B4    		& C  			& M 		      	& R			& O       			\\
                \midrule
                SCN \citep{DBLP:conf/cvpr/GanGHPTGCD17}  	          		& 42.1      	  	& 48.3		& 28.7		& 61.6		& 0.9152			\\
                SCN\footnotemark[1]                                   				& 44.3          	& 51.5            	& 29.7		& 63.0		& 0.9550  			\\
                SCN\footnotemark[2]                                 					& 44.4          	& 50.4            	& 29.7		& 63.0		& 0.9506  			\\
                \midrule
                ECO \citep{zolfaghari2018eco}         				      		& 43.0   		& 49.8  		& 28.9		& 62.1		& 0.9304			\\
                ECO\footnotemark[1]                                 	 				& 44.4        	& 51.6            	& 29.5      		& 63.1		& 0.9548  			\\
                ECO\footnotemark[2]                                 					& 44.5        	& 50.6            	& 29.6      		& 63.1		& 0.9516  			\\
                \midrule
                SAM-SS \citep{10.3389/frobt.2020.475767}				& 43.8      		& 51.4    		& 28.9      		& 62.4		& 0.9431			\\
                SAM-SS\footnotemark[1]								& 44.9      		& 52.3            	& 29.5        	& 63.2		& 0.9610  			\\
                SAM-SS\footnotemark[2]								& 45.0      		& 51.1            	& 29.5        	& 63.2		& 0.9561  			\\
                \midrule
                VNS-GRU \citep{chen2020delving}				              	& 45.3		& 53.0		& 29.9		& 63.4		& 0.9704  			\\
                VNS-GRU\footnotemark[1]           					        	& \textbf{46.9} 	& \textbf{55.1}   &\textbf{30.8}  	&\textbf{64.5}	& \textbf{1.0000}  	\\
                VNS-GRU\footnotemark[2]   					        		& 46.6          	& 52.2            	& 30.6          	& 64.3		& 0.9828  			\\
                \bottomrule
            \end{tabular*}
            \footnotetext[1]{The model was trained on the cleaned training set and the metrics were calculated on the original test set.}
            \footnotetext[2]{The model was trained on the cleaned training set and the metrics were calculated on the cleaned test set.}
        \end{minipage}
    \end{center}
\end{table*}

\subsection{Comparison between the Original/Cleaned MSR-VTT Datasets}\label{experiment:comparison}
In Table \ref{table:performanceMSRVTTNewOld}, a model name without any superscript indicates that
the model was trained on the original training set and the metrics were calculated on the original test set.
We had four observations.
First, the models trained on the cleaned training set achieved higher scores of metrics than the models trained on the original training set, even though the metrics were calculated on the original test set. 
For instance, VNS-GRU$^1$ \citep{chen2020delving} improves over VNS-GRU by 1.6\% on BLEU-4, by 2.1\% on CIDEr, by 0.9\% on METEOR and by 1.1\% on ROUGE-L.
Second, the models trained on the cleaned training set and tested on the cleaned test set achieved higher scores of metrics than the models trained on the original training set and tested on the original test set. 
For instance, VNS-GRU$^2$ \citep{chen2020delving} improves over VNS-GRU by 1.3\% on BLEU-4, by 0.7\% on METEOR and by 0.9\% on ROUGE-L.
Third, the scores of VNS-GRU$^2$ were slightly lower than the scores of VNS-GRU$^1$.
We attributed this to the increase of annotation diversity in the cleaned test set.
\begin{figure*}[htb]
    \centering
    \includegraphics[width=.9\textwidth]{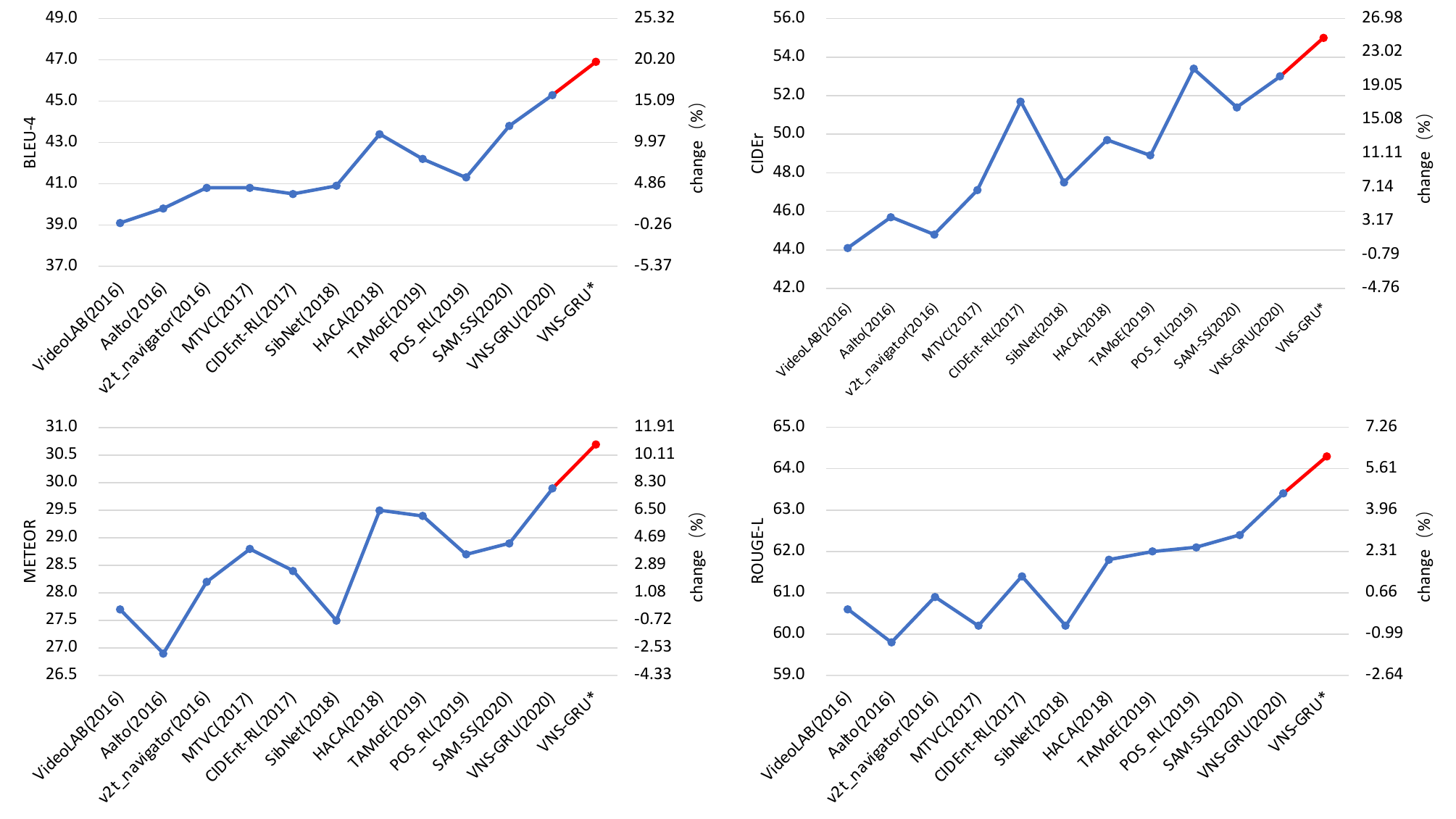}
    \caption{The performance of typical models on the MSR-VTT dataset during 2016 and 2020.
    The models include VideoLAB, Aalto, v2t\_navigator,
    MTVC \citep{DBLP:conf/acl/PasunuruB17}, CIDEnt-RL \citep{pasunuru-bansal-2017-reinforced},
    SibNet \citep{liu-sheng-2018-sibnet}, HACA \citep{wang2018watch}, TAMoE \citep{wang2019learning},
    SAM-SS \citep{10.3389/frobt.2020.475767} and POS\_RL \citep{Wang_2019_ICCV} and VNS-GRU \citep{chen2020delving}.
    The first three models are from ACM Multimedia MSR-VTT Challenge 2016 \citep{acm/2016/Online}. VideoLAB was used as the baseline (0\% change).}
    \label{fig:msrvttImprovement}
\end{figure*}
Fourth, the improvement in overall score \eqref{eq:overall} is comparable to or larger than the SOTA methods in recently years. 
The overall score of the model trained on the cleaned dataset is higher than the one trained on the original dataset by nearly 3.0\%, 1.8\%, 2.4\%, 4.0\% for VNS-GRU, SAM-SS, ECO, and SCN, respectively. 
For comparison, a new method presented in \citep{wang2018watch} called Hierarchically aligned cross-modal attention (HACA) framework improves the overall score of the previous state-of-the-art model CIDEnt-RL by 1.8\% (the overall score is calculated according to \eqref{eq:overall}) based on Table 1 in \citep{wang2018watch}. 
A new method Retrieve-Copy-Generate network presented in  \citep{zhang2021open} improves the overall score of the previous state-of-the-art model ORG-TRL by 0.75\% according to Table 6 in \citep{zhang2021open}.

The scores of BLEU-4, CIDEr, METEOR and ROUGE-L of popular video captioning models, proposed in recent years, are plotted in Fig. \ref{fig:msrvttImprovement}.
One of the earliest models on the MSR-VTT dataset,
VideoLAB, from ACM Multimedia MSR-VTT Challenge 2016 \citep{acm/2016/Online}, was used as the baseline,
and all other models were compared with it.
Then the relative changes of other models in percentage can be inferred on the right vertical axes in Fig. \ref{fig:msrvttImprovement}.
By training on the cleaned training set, one of the state-of-the-art models, VNS-GRU was improved from 15.9\% to 19.9\% on BLEU-4, from 20.2\% to 24.7\% on CIDEr, from 7.9\% to 10.8\% on METEOR,
from 4.6\% to 6.1\% on ROUGE-L, compared with the results obtained by the same model trained on the original training set.
From the figure, it is seen that the relative improvements brought by annotation cleaning were non-negligible.

\subsection{Ablation Study}\label{experiment:effectiveness}
To analyze the utility of each step in data cleaning,
we compared the performances of the model VNS-GRU \citep{chen2020delving} on the original and cleaned test sets in Tables \ref{table:steps1} and \ref{table:steps2},
trained on the training set cleaned by Step I (Section \ref{step1}),
Step II (Section \ref{step2}), Step III (Section \ref{step3}), Step IV (Section \ref{step4}), accumulatively.

    \begin{table}[htb]
        \begin{center}
        	\begin{minipage}[c]{\linewidth}
                \caption{Results on the origin test set }
                \label{table:steps1}
                \begin{tabular*}{\textwidth}{@{\extracolsep{\fill}}ccccccccc@{\extracolsep{\fill}}}
                \toprule
                    I \footnotemark[1]	& II					& III						& IV					& B4			& C 				& M 				& R		& O				\\
                \midrule
                $\times$			& $\times$			& $\times$				& $\times$			& 	45.3		&	53.0			&	29.9			&	63.4	&	0.9678		\\
                $\surd$			& $\times$			& $\times$				& $\times$			& 	47.1		&	54.4			&	30.4			&	64.0	&	0.9901		\\
                $\surd$			& $\surd$				& $\times$				& $\times$			& 	47.3		&	53.9			&	30.2			&	64.1	&	0.9876		\\
                $\surd$			& $\surd$				& $\surd$					& $\times$			& 	47.4		&	55.0			&	30.7			&	64.2	&	0.9975		\\
                $\surd$			& $\surd$				& $\surd$					& $\surd$				& 	46.9		&	55.1			&	30.8			&	64.5	&	0.9974		\\	
                \bottomrule		
                \end{tabular*}
                \footnotetext[1]{ The model was trained on the training set with data cleaning steps I, II, III and IV taken one by one. }
            \end{minipage}
        \end{center}
    \end{table}

As shown in Tables \ref{table:steps1} and \ref{table:steps2}, Step I brought improvements in all the metrics since it reduced the number of irregular words and phrases, which contain special characters.
After Step II, the four metrics remained similar to those after Step I when measured on the original test set (Table \ref{table:steps1}), 
but the metrics were improved when measured on the cleaned test set (Table \ref{table:steps2}).
After Step III, all metrics except METEOR increased in the both cases.
The METEOR value slightly decreased when measured on the cleaned test set (Table \ref{table:steps2}).
After the last step, almost all metrics were further improved, except BLEU-4.
If we focus on the performance of the model measured on the cleaned test set (Table \ref{table:steps2}), we found that the overall score was improved after each step.
These results suggest that all steps are necessary for cleaning the annotations.

    \begin{table}[htb]
        \begin{center}
        	\begin{minipage}{\linewidth}
                \caption{Results on the cleaned test set. The model was trained on the training set with data cleaning steps I, II, III and IV taken one by one }
                \label{table:steps2}
                \begin{tabular*}{\textwidth}{@{\extracolsep{\fill}}ccccccccc@{\extracolsep{\fill}}}
                    \toprule
                        I					& II				& III				& IV				& B4			& C 			& M 				& R  		&	O		\\
                    \midrule
                    $\times$				& $\times$		& $\times$		& $\times$		& 	44.5		&	49.8		&	29.7			&	63.0	&	0.9598	\\
                    $\surd$				& $\times$		& $\times$		& $\times$		& 	46.9		&	50.9		&	30.2			&	63.8	&	0.9849	\\
                    $\surd$				& $\surd$			& $\times$		& $\times$		& 	46.9		&	51.4		&	30.3			&	63.9	&	0.9885	\\
                    $\surd$				& $\surd$			& $\surd$			& $\times$		& 	47.6		&	51.7		&	30.2			&	64.1	&	0.9936	\\
                    $\surd$				& $\surd$			& $\surd$			& $\surd$			& 	46.6		&	52.2		&	30.6			&	64.3	&	0.9947	\\	
                    \bottomrule		
                \end{tabular*}
            \end{minipage}
        \end{center}
    \end{table}

\section{Human Evaluation}

    \begin{figure}[htb]
        \centering
        \includegraphics[width=0.9\linewidth]{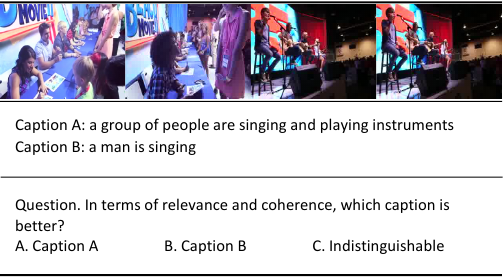}
        \caption{An example question in the human evaluation experiment. Captions A and B were generated by VNS-GRU or VNS-GRU* }
        \label{fig:question-example}
    \end{figure}
    
It is well-known that the metrics including BLEU-4 \citep{papineni-etal-2002-bleu}, CIDEr \citep{cider-2015}, METEOR \citep{banerjee-lavie-2005-meteor}, ROUGE-L \citep{lin-2004-rouge} do not fully reflect the quality of the video captioning results.
We then conducted a human evaluation study.
We recruited 17 people (11 male and 6 female, ages between 20 and 35) with normal or corrected-to-normal vision to do this experiment.
The subjects were mainly from Tsinghua University, Beijing, China.
All subjects had at least college level English.
This study was approved by the Department of Psychology Ethics Committee, Tsinghua University, Beijing, China.

The subjects watched video clips from the MSR-VTT dataset and compared the results of VNS-GRU trained on the original and cleaned annotations of the dataset (Figure \ref{fig:question-example}).
The subjects were instructed to compare the results based on two criteria:
    \begin{enumerate}
        \item relevance, the match between the contents of the video clip and the caption;
        \item coherence, the language fluency and grammatical correctness in the caption.
    \end{enumerate}
For each video clip, there were three options: (A) Caption A is better; (B) Caption B is better; and (C) Indistinguishable. The two captions were generated by VNS-GRU or VNS-GRU*, which were trained on the original and cleaned annotations of the dataset, respectively. The subjects needed to choose one and only one of three options. A total of 30 video clips were randomly sampled from the test set and presented to all subjects in an fixed order. Every subject completed the experiment within half an hour.
    
    \begin{figure}[htb]
      \centering
      \includegraphics[width=0.9\linewidth]{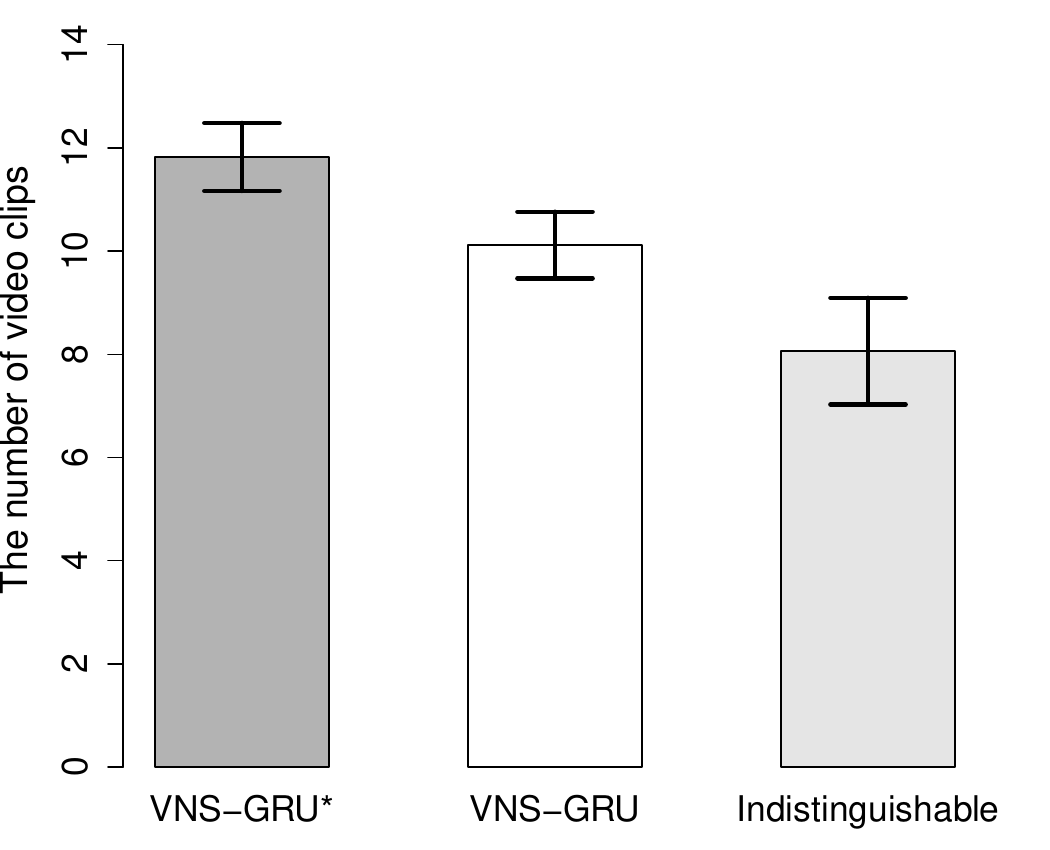}
      \caption{Human evaluation results. ``VNS-GRU$^*$'', ``VNS-GRU'' and ``Indistinguishable'' denote the numbers of videos which the subjects
      voted for ``VNS-GRU$^*$ is better than VNS-GRU'',
      ``VNS-GRU is better than VNS-GRU$^*$'' and
      ``They are indistinguishable'', respectively. Error bars are standard deviations.
      The p-value between ``VNS-GRU$^*$'' and ``VNS-GRU'' is 0.02}
      \label{fig:votePerUser}
    \end{figure}
    
We noted down the number of votes for VNS-GRU, VNS-GRU* and Indistinguishable for every subject and calculated the average over all subjects (Figure \ref{fig:votePerUser}). 
On average, for 11.8 video clips the subjects voted for ``VNS-GRU* is better'' and for 10.1 video clips the subjects voted for ``VNS-GRU is better''. 
The one-sided student t-test indicated that VNS-GRU* performed better than VNS-GRU ($p=0.02, n=17$). On average, for 8.1 videos the subjects could not distinguish the quality of the results.

These results suggested that annotation cleaning could boost the quality of the generated captions by video captioning models from subjective evaluation of human. 
Note that the difference in human evaluation between the original dataset and cleaned dataset is significant, but not very large. 
It might be due to the fact that many of the human subjects are not native English speakers and they might have relatively insufficient ability to judge the difference in quality of the generated sentences.

\section{Conclusion}
The MSR-VTT dataset is a widely used dataset in the areas of video captioning and video retrieval.
Thousands of problems were found in its annotations, and many of them were obvious mistakes.  
We inspected the influence of these problems on the results of video captioning models. 
By four steps of data cleaning, we removed or corrected sentences to resolve these problems, and compared the results of several popular video captioning models.
The models trained on the cleaned dataset generated better captions than the models trained on the original dataset measured by both objective metrics and subjective evaluations.
In particular, trained on the cleaned dataset, VNS-GRU achieved better results with improvement of at least 0.9\% compared to the baseline.
This cleaned dataset is recommended for developing new video captioning models in the future. 
And the proposed method can also be applied to other datasets, including NLP-only datasets, to help model training. 

\section*{Acknowledgments}
The authors would like to thank Han Liu and Huiran Yu for insightful discussions. This work was supported by the National Natural Science Foundation of China under Grant 62061136001,  Grant U19B2034 and Grant 61620106010, and by the deutsche Forschungsgemeinschaft (DFG, German Research Foundation) -- TRR 169/A6.

\bibliographystyle{model2-names}
\bibliography{refs}

\end{document}